\documentclass[runningheads]{llncs}
\usepackage{graphicx}
\usepackage{multirow}
\usepackage{amsmath}
\usepackage{amssymb}

\renewcommand{\vec}[1]{\mathbf{#1}}
\renewcommand{\matrix}[1]{\mathbf{#1}}

\newcommand{\es}{e_\mathrm{s}}
\newcommand{\relq}{r_\mathrm{q}}
\newcommand{\eo}{e_\mathrm{o}}

\begin{document}
\title{Explainable Link Prediction for Emerging Entities in Knowledge Graphs}
\author{Rajarshi Bhowmik\inst{1} \and
Gerard de Melo\inst{2}}
\authorrunning{R.\ Bhowmik and G.\ de Melo}
\institute{Rutgers University--New Brunswick\\Piscataway, New Jersey, USA \and 
Hasso Plattner Institute, University of Potsdam\\Potsdam, Germany\\
\email{rajarshi.bhowmik@rutgers.edu; gdm@demelo.org}}
\maketitle              %
\begin{abstract}
Despite their large-scale coverage, cross-domain knowledge graphs invariably suffer from inherent incompleteness and sparsity. Link prediction can alleviate this by inferring a target entity, given a source entity and a query relation. Recent embedding-based approaches operate in an uninterpretable latent semantic vector space of entities and relations, while path-based approaches operate in the symbolic space, making the inference process explainable. However, these approaches typically consider static snapshots of the knowledge graphs, severely restricting their applicability for evolving knowledge graphs with newly emerging entities. To overcome this issue, we propose an inductive representation learning framework that is able to learn representations of previously unseen entities. Our method finds reasoning paths between source and target entities, thereby making the link prediction for unseen entities interpretable and providing support evidence for the inferred link.

\end{abstract}

\section{Introduction}
Recent years have seen a surge in the usage of large-scale cross-domain knowledge graphs \cite{KnowledgeGraphs2020}
for various 
tasks, including factoid question answering, fact-based dialogue engines, and information retrieval \cite{BhowmikDeMelo2019EntityDescriptionsCopyModel}. Knowledge graphs serve as a source of background factual knowledge for a wide range of applications \cite{deMelo2018KnowledgeGraphs}. 
For example, Google's knowledge graph is tightly integrated into its search engine, while Apple adopted Wikidata as a source of background knowledge for its virtual assistant Siri. 
Many such applications deal with 
queries that can be transformed to a structured relational query of the form $(\es, \relq, ?)$, where $\es$ is the source entity and $\relq$ is the query relation. 
For example, the query ``\emph{Who is the director of World Health Organization?}'' can be mapped to the structured query \emph{(World Health Organization, director, ?)} while executing it on a knowledge graph.
Unfortunately, due to the inherent sparsity and incompleteness of knowledge graphs, answers to many such queries cannot be fetched directly from the existing data, but instead need to be inferred indirectly.

Furthermore, with the ever-increasing volume of the knowledge graphs, the number of emerging entities also increases. Many of these emerging entities have a small number of known facts at the time they are integrated into the knowledge graphs. Therefore, their connectivity to pre-existing entities in the knowledge graph is often too sparse. 

\begin{figure}
    \centering
    \includegraphics[height=1.92in, width=0.56\linewidth]{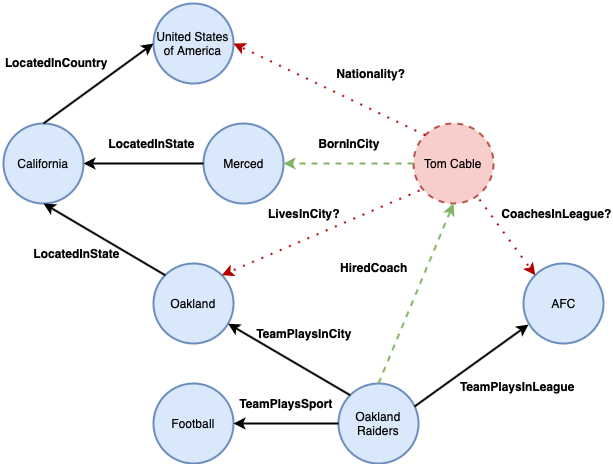}
    \caption{A subgraph of NELL with \emph{Tom Cable} as an emerging entity. The solid-lined circles and arrows represent the existing entities and relations. The dashed-lined circles and arrows denote an emerging entity and some of its known relationships to other existing entities. The unknown relationships that need to be inferred through inductive representation learning and explainable reasoning are shown as dotted arrows.}
    \label{fig:example_kg}
\end{figure}

In recent years, embedding-based models \cite{Nguyen17a} have widely been adopted to infer missing relationships in a knowledge graph. In such embedding-based models, distributed vector representations of entities and relations in the knowledge graph are used to learn a scoring function $f(\es, \relq, \eo)$ in a latent embedding space to determine the plausibility of inferring a new fact.
However, these models are lacking in terms of the interpretability and explainability of the decisions they make. One does not obtain any clear explanation of why a specific inference is warranted. For example, from the embeddings of facts \emph{(A, born\_in, California)} and \emph{(California located\_in, US)}, the fact \emph{(A, born\_in, US)} could be deduced. But logical composition steps like this one are learned implicitly by knowledge graph embeddings. This means that this approach cannot offer such logical inference paths as support evidence for an answer. 

In contrast, path-based reasoning approaches operate in the symbolic space of entities and relations, leveraging the symbolic compositionality of the knowledge graph relations, thus making the inference process explainable. This means that the user can inspect the inference path, consisting of  existing edges in the knowledge graph, as support evidence. To this end, purely symbolic and fast rule-mining systems, e.g., PRA \cite{Lao:2011}, AMIE+ \cite{AMIE+:2015}, and AnyBURL \cite{AnyBURL:2019:Meilicke} may attain a level of performance that is comparable to embedding-based methods, but neglect many of the statistical signals exploited by the latter. To leverage the advantages of both path-based and embedding-based models, some neural-symbolic approaches \cite{Gardner:2013,Gardner:2014,Neelakantan:2015,Guu:2015,RuleS:2018:Ho} have as well been proposed.
Some recent path-based reasoning approaches \cite{Das:2017,Lin:2018a} formulate the path-finding problem as a Partially Observable Markov Decision Process (POMDP), in which the model learns a policy to find an inference path from the source entity to the target entity using REINFORCE \cite{Williams:1992}, a policy gradient based reinforcement learning algorithm. 

However, most of these approaches are studied with static snapshots of the knowledge graphs, thus severely restricting their applicability for a dynamically evolving knowledge graph with many emerging entities. Except for the purely symbolic rule-mining systems mentioned above, most existing approaches that depend on learning latent representations of entities require that all entities are present during training. Therefore, these models are incapable of learning representations of arbitrary newly emerging entities not seen during training. Some recent approaches such as HyTE \cite{dasgupta-etal-2018-hyte} and DyRep \cite{Trivedi2019DyRepLR} have considered dynamically evolving temporal knowledge graphs. However, similar to embedding-based models, these models are not explainable.

To overcome this issue, we propose a joint framework for representation learning and reasoning in knowledge graphs that aims at achieving inductive node representation learning capabilities applicable to a dynamic knowledge graph with many emerging entities while preserving the unique advantage of the path-based approaches in terms of explainability. For inductive node representation learning, we propose a variant of \emph{Graph Transformer} encoder \cite{koncel-kedziorski-etal-2019:GraphTransformer:ACL2019,Yun:GraphTransformer:NIPS2019} that aggregates neighborhood information based on its relevance to the query relation. Furthermore, we use policy gradient-based reinforcement learning (REINFORCE) to decode a reasoning path to the answer entity. 
We hypothesize that the inductively learned embeddings provide prior semantic knowledge about the underlying knowledge environment to the reinforcement learning agent.

We summarize the contributions of this paper as follows: (1) We introduce a joint framework for inductive representation learning and explainable reasoning that is capable of learning representations for unseen emerging entities during inference by leveraging only a small number of known connections to the other pre-existing entities in the knowledge graph. Our approach can not only infer new connections between an emerging entity and any other pre-existing entity in the knowledge graph, but also provides an explainable reasoning path as support evidence for the inference. (2) We introduce new train/development/test set splits of existing knowledge graph completion benchmark datasets that are appropriate for inductive representation learning and reasoning.

\section{Related Work}
\subsection{Embedding-based Methods} 
Knowledge graph completion can be viewed as an instance of the more general problem of link prediction in a graph \cite{WangEtAl2017LinkPredictionNFGEG}. Due to advances in representation learning, embedding-based methods have become the most popular approach. Such methods learn $d$-dimensional distributed vector representations of entities and relations in a knowledge graph. To this end, the translation embedding model TransE \cite{Bordes:2013} learns the embedding of a relation as a simple translation vector from the source entity to the target entity such that $\vec{e}_s + \vec{e}_r \approx \vec{e}_o$. Its variants, e.g., TransH \cite{TransH:Wang:2014:AAAI}, TransR \cite{TransR:Lin:2015:AAAI}, TransD \cite{TransD:Ji:2015:ACL} consider similar objectives.
Tri-linear models such as DistMult \cite{Yang:2014}, along with its counterpart ComplEx \cite{Trouillon:2016} in the complex embedding space, use a multiplicative scoring function $f(s, r, o) = \vec{e}_s^\intercal \mathbf{W}_r \vec{e}_o$, where $\textbf{W}_r$ is a diagonal matrix representing the embedding of relation $r$.
Convolutional neural network models such as ConvE \cite{Dettmers:2018} and ConvKB \cite{ConvKB:Nguyen:2018:NAACL} apply convolutional kernels over entity and relation embeddings to capture the interactions among them across different dimensions.
These models obtain state-of-the-art results on the benchmark KBC datasets. However, none of the above-mentioned approaches deliver the full reasoning paths that license specific multi-hop inferences, and hence they either do not support multi-hop inference or do not support it in an interpretable manner. Moreover, these approaches assume a static snapshot of the knowledge graph to train the models and are not straightforwardly extensible to inductive representation learning with previously unseen entities.

\subsection{Path-based Methods}
An alternative stream of research has explored means of identifying specific paths of inference, which is the task we consider in this paper. To this end, the Path Ranking Algorithm (PRA) \cite{Lao:2011} uses random walks with restarts for multi-hop reasoning. Following PRA, other approaches \cite{Gardner:2013,Gardner:2014,Neelakantan:2015,Guu:2015} also leverage random walk based inference. However, the reasoning paths that these methods follow are gathered by random walks independently of the query relation. 

Recent approaches have instead adopted policy gradient based reinforcement learning for a more focused exploration of reasoning paths. Policy gradient based models such as DeepPath \cite{Xiong:2017a}, MINERVA \cite{Das:2017}, MINERVA with Reward Shaping and Action Dropout \cite{Lin:2018a}, and M-Walk \cite{Shen:2018:M-Walk} formulate the KG reasoning task as a Partially Observable Markov Decision Process and learn a policy conditioned on the query relation. Such reasoning techniques have also been invoked for explainable recommendation \cite{XianFuMuthukrishnanDeMeloZhang2019KGRecommendations,FuXianGaoZhaoHuangGeXuGengShahZhangDeMelo2020FairRecommendation,XianFuZhaoGeChenHuangGengQinDeMeloMuthukrishnanZhang2020CAFE} and explainable dialogue systems \cite{YangXinyuWangZhangDeMelo2020ExplainableDialogueIntent}.
Although the inference paths are explainable in these models (if reward shaping is omitted), there may be a substantial performance gap in comparison with embedding-based models. 

Another sub-category of path-based methods, e.g., AMIE+ \cite{AMIE+:2015}, AnyBURL \cite{AnyBURL:2019:Meilicke}, and RuleS \cite{RuleS:2018:Ho} proceed by mining Horn rules from the the existing knowledge graphs for link prediction. The body of a Horn rule provides the reasoning path. Although these approaches are capable of fast rule mining and can easily be applied to unseen emerging entities, the quality of the learned rules are affected by the sparsity of the knowledge graph.

\subsection{Graph Convolution-based Methods}
Graph Convolution Networks (GCNs) can be used for node classification in a homogeneous graph \cite{GCN:Kipf:2017:ICLR}. They are an instance of Message Passing Neural Networks (MPNN), in which the node representations are learned by aggregating information from the nodes' local neighborhood. GraphSAGE \cite{GraphSAGE:Hamilton:2017:NIPS} attempts to reduce the memory footprint of GCN by random sampling of the neighborhood. Graph Attention Networks (GAT) \cite{GAT:Velickovic:2018:ICLR} are a variant of GCN that learn node representations as weighted averages of the neighborhood information. However, GCN and its variants such as GAT and GraphSAGE are not directly applicable for link prediction in knowledge graphs, as they ignore the edge (relation) information for obtaining the node embeddings. To alleviate this issue, R-GCNs operate on relational multi-graphs \cite{Schlichtkrull:2018}, but, similar to GCNs, R-GCNs also need all nodes of the graphs to be present in memory and therefore are not scalable to large-scale knowledge graphs. Hamaguchi et al.~\cite{Hamaguchi:2017:IJCAI} proposed a model for computing representations for out-of-KG entities using graph neural networks. The recent models such as SACN \cite{SACN:Shang:2019:AAAI} and CompGCN \cite{CompGCN:2020:Vashishth} leverage the graph structure by inductively learning representations for edges (relations) and nodes (entities). However, unlike our model, these methods are not explainable.

\section{Model}
\begin{figure*}
\centering
\includegraphics[height=2.24in, width=0.8\linewidth]{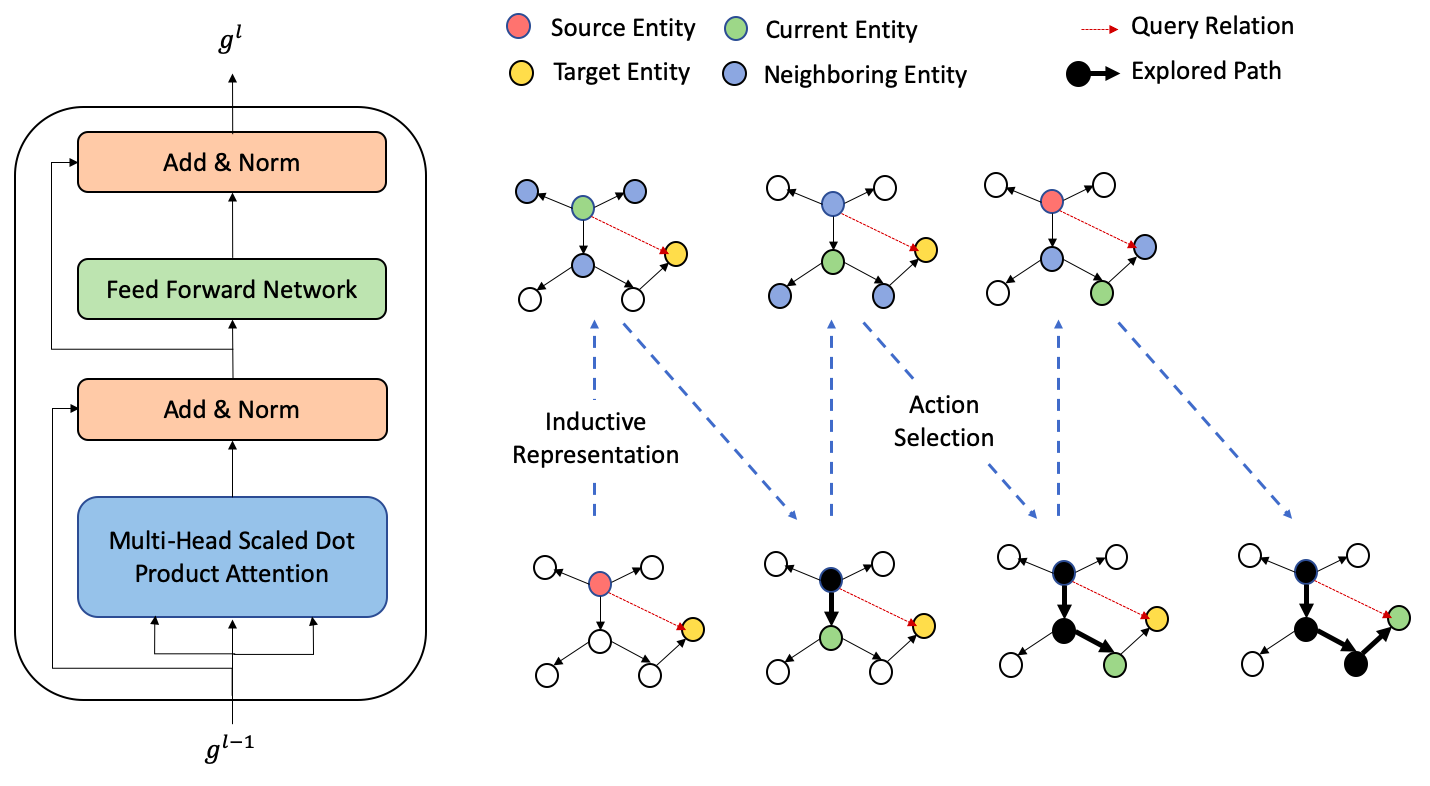}
\caption{A schematic diagram of a Graph Transformer block, along with an illustration of the workflow of our model, demonstrating successive applications of inductive node representation learning and action selection to find a reasoning path.}
\label{fig:encoder}
\end{figure*}

Our model consists of two modules that are subject to joint end-to-end training. The encoder module learns inductive entity embeddings while accounting for the query relation and the local neighborhood of an entity (Section \ref{sec:graph-transformer}). The decoder module operates on this learned embedding space of entities and relations. By leveraging the embeddings of the source entity and the query relation, the decoder module infers a reasoning path to the target entity using policy gradient-based reinforcement learning (Section \ref{sec:pg}). Before describing these components in more detail, Section \ref{sec:problem-statement} first provides preliminary definitions.

\subsection{Problem Statement}\label{sec:problem-statement}
Formally, we consider knowledge graphs $\mathcal{G}(\mathcal{E}, \mathcal{R}, \mathcal{F})$ defined as directed multi-graphs such that each node $e \in \mathcal{E}$ represents an entity, each $r \in \mathcal{R}$ represents a unique relation, and each directed edge $(\es, r, \eo) \in \mathcal{F}$ represents a fact about the subject entity $\es$.

Given a structured relational query $(\es, \relq, ?)$, where $\es$ is the source entity, $\relq$ is the query relation, and $(\es, \relq, \eo) \notin \mathcal{F}$, the goal is to find a set of plausible answer entities $\{\eo\}$ by navigating paths through the existing entities and relations in $\mathcal{G}$ leading to answer entities. Note that, unlike previous methods that consider transductive settings with a static snapshot of the knowledge graph, we allow for dynamic knowledge graphs, where $\es$ may be an emerging entity, and therefore, previously unseen. Moreover, while embedding-based methods only deliver candidate answer entities, we here also seek the actual paths, i.e., sequences of nodes and edges for better interpretability.\footnote{From here onwards, we will use the terms \emph{node} and \emph{entity}, as well as \emph{edge} and \emph{relation(ship)} interchangeably.}

\subsection{Graph Transformer for Inductive Representation Learning}\label{sec:graph-transformer}
The state-of-the-art embedding based models either focus on learning entity embeddings by using only the query relations, ignoring the subject entity's neighborhood, or use message passing neural networks to learn embeddings conditioned on neighboring entities and relations while being oblivious of the query relation.
However, we observe that in many cases a new fact can be inferred by using another existing fact. For example, the fact \emph{(PersonX, Place of Birth, Y)} can often help to answer to the query \emph{(PersonX, Nationality, ?)}. Motivated by this observation, we propose a Graph Transformer architecture that learns the embedding of the source entity by iterative aggregation of neighborhood information (messages) that are weighted by their relevance to the query relation. To learn the relevance weights, our Graph Transformer model deploys \emph{multi-head scaled dot product attention}, also known as \emph{self-attention} \cite{Vaswani:2017:NIPS}.  

Formally, we denote the local neighborhood for each entity $e_i \in \mathcal{E}$ as $\mathcal{N}_i$ such that $\mathcal{N}_i = \{ e_j \mid e_j \in \mathcal{E} \wedge (e_i, r, e_j) \in \mathcal{F} \wedge r \in \mathcal{R}_{ij} \}$, where $\mathcal{R}_{ij}$ is the set of relations between entities $e_i$ and $e_j$.

Each neighboring entity $e_j \in \mathcal{N}_i$ connected to $e_i$ by a relation $r$ sends in a message to entity $e_i$. The message $\vec{m}_{ijr}$ is a linear transformation of the fact $(e_i, r, e_j)$ followed by the application of a non-linear function, specifically, the leaky rectified linear unit (LeakyReLU) function with a negative slope of $0.01$. Formally, 
\begin{equation}
    \vec{m}_{ijr} = \mathrm{LeakyReLU}(\matrix{W}_f [\vec{e_i}; \vec{r}; \vec{e_j}]),
\end{equation}
where $\matrix{W}_f \in \mathbb{R}^{d \times 3d}$ is a shared parameter for the linear transformation and [;] is the concatenation operator.

To compute an attention head, our model performs linear projections of the query relation $\relq$, the neighborhood relations $r \in \mathcal{R}_{ij}$, and the neighborhood messages $\vec{m}_{ijr}$ to construct queries $Q$, keys $K$, and values $V$, respectively, such that 
$Q = \matrix{W}_Q \vec{\relq} $, $K =\matrix{W}_K \vec{r}$, and $V = \matrix{W}_V \vec{m}_{ijr}$,
where $\matrix{W}_Q, \matrix{W}_K, \matrix{W}_V \in \mathbb{R}^{d' \times d}$ are learnable parameters.

Next, we use the queries $Q$ to perform a dot-product attention over the keys $K$. Formally,
\begin{equation}
    \alpha_{ijr} = \frac{\mathrm{exp}((\matrix{W}_Q \vec{\relq})^ \intercal (\matrix{W}_K \vec{r}))}{\displaystyle \sum_{z \in \mathcal{N}_i} \sum_{r' \in \mathcal{R}_{ij}} \mathrm{exp}((\matrix{W}_Q \vec{\relq})^ \intercal (\matrix{W}_K \vec{r'}))}
\end{equation}
We adopt the common procedure of scaling the dot products of $Q$ and $K$ by a factor of $\frac{1}{\sqrt{d'}}$ \cite{Vaswani:2017:NIPS}.

The attention weights are then used to aggregate the neighborhood messages. Note that \emph{self-attention} deploys multiple attention heads, each having its own query, key, and value projectors. The aggregated messages from $N$ attention heads are concatenated and added to the initial embedding $\vec{e_i}$ through a residual connection to obtain new intermediate representation 
\begin{equation}
\vec {\hat{e}_i}= \vec{e_i} + \Vert_{n=1}^N \left(\sum_{j \in \mathcal{N}_i} \sum_{r \in \mathcal{R}_{ij}} \alpha^{n}_{ijr}  \matrix{W}_V^n \vec{m}_{ijr}\right),
\end{equation}
where $\Vert$ is the concatenation operator.

Layer normalization (LN) \cite{LayerNorm:2016:Ba} is applied to the intermediate representation  $\vec{\hat{e}}_i$, followed by a fully connected two-layer feed forward network (FFN) with a non-linear activation (ReLU) in between. Finally, the output of the feed forward network is added to the intermediate representation through another residual connection. The resulting embedding is again layer normalized to obtain the new representation $\vec{g_i}^l$ for $e_i$. Formally,
\begin{equation}
    \vec{g_i}^l = \mathrm{LN}(\mathrm{FFN}(\mathrm{LN}(\vec {\hat{e}_i})) + \mathrm{LN}({\vec {\hat{e}_i}}))
\end{equation}

This pipeline is called a \emph{Transformer block}. Figure \ref{fig:encoder} represents a schematic diagram of a Transformer block in Graph Transformers. We stack $L$ layers of Transformer blocks to obtain the final embedding $\vec{g_i}^L$ for $e_i$.

\subsection{Policy Gradient for Explainable Reasoning}
\label{sec:pg}
To infer the answer entity, we could leverage the entity representations obtained by the Graph Transformers. However, our goal is not only to infer the answer entity, but to find a symbolic reasoning path to support the inference.
Following previous work \cite{Das:2017,Lin:2018a}, we formulate the reasoning task as a finite horizon, deterministic partially observable Markov Decision Process (POMDP). A knowledge graph can be seen as a partially observable environment with out-going relations at each entity node corresponding to a set of discrete actions that an agent can explore to reach the target answer from the source entity.
\paragraph{Knowledge Graph Environment.} Formally, a Markov Decision Process is defined by a 4-tuple $(\mathcal{S}, \mathcal{A}, \mathcal{P}, \mathcal{R})$, where $\mathcal{S}$ is a finite set of states, $\mathcal{A}$ is a finite set of actions, $\mathcal{P}$ captures state transition probabilities, and $\mathcal{R}$ is the reward function.
In a knowledge graph environment, the state space is defined as a set of tuples $s_t = (e_t, \relq) \in \mathcal{S}$, where $e_t$ is an entity node in the knowledge graph, 
and $\relq$ is the query relation. The action space $A_t \in \mathcal{A}$ for a state $s_t$ is defined as the set of outgoing edges from entity node $e_t$ in the knowledge graph. Formally, $A_t = \{(r_{t+1}, s_{t+1}) \mid (e_t,r_{t+1}, s_{t+1}) \in \mathcal{G}\}$. Since state transitions in a KG environment are deterministic, the transition probabilities $P(s_{t+1} \mid s_t, a_t) = 1\,\, \forall P \in \mathcal{P}$. The agent receives a terminal reward of $1$ if it arrives at the correct answer entity at the end. 
\paragraph{Graph Search Policy.}
To find a plausible path to the answer entity, the model must have a policy to choose the most promising action at each state. Note that in the KG environment, the decision of choosing the next action is not only dependent on the current state, but also on the sequence of observations and actions taken so far in the path. We use a multi-layer LSTM as a sequence encoder to encode the path history.

Formally, each state $s_t$ is represented by a vector $\vec{s}_t = [\vec{e}_t; \vec{r}_q] \in \mathbb{R}^{2d}$ and each possible action $a_t \in A_t$ is represented by $\vec{a}_t = [\vec{e}_{t+1}; \vec{r}_{t+1}] \in \mathbb{R}^{2d}$, where $\vec{e}_t$, $\vec{e}_{t+1} \in \mathbb{R}^d$ are the embeddings of %
the entity nodes at timesteps $t$ and $t+1$, respectively, that are obtained from Graph Transformer encoders. $\vec{r}_{t+1} \in \mathbb{R}^d$ is the embedding of an out-going relation from entity $e_t$, and $\vec{r}_q \in \mathbb{R}^d$ corresponds to the embedding of the query relation $\relq$. Each of these embeddings is also obtained from the Graph Transformer encoder. The path history is encoded as $\vec{h}_{t} = \mathrm{LSTM}(\vec{h}_{t-1}, \vec{a}_{t-1})$.
Given the embedded action space $\vec{A}_t \in \mathbb{R}^{2 |A_t|}$, i.e., the stacked embeddings of actions $a_t \in A_t$, and the path history $\vec{h}_t$, we define the parameterized policy as:
\begin{equation*}
    \pi_{\theta}(a_t \mid s_t) = \mathrm{Softmax}(\matrix{A}_t(\matrix{W_2}\mathrm{ReLU}(\matrix{W_1}[\vec{h}_t;\vec{e}_t; \vec{r}_q])))
\end{equation*}

\paragraph{Policy Optimization.}
The policy network is trained to maximize the expected reward for all $(\es, \relq, \eo)$ triples in the training sub-graph. The agent learns an optimal policy $\pi_\theta$ by exploring a state space of all possible actions. The objective of the agent is to take actions to maximize the expected end reward.
Formally:
\begin{equation}
     J(\theta) = \mathbb{E}_{(\es, \relq, \eo)}\left[ \mathbb{E}_{a_1, ..., a_{T-1} \sim \pi_{\theta}}[R(s_T|\es, \relq)]\right]
\end{equation}

Since policy gradient uses gradient-based optimization techniques, the estimated gradient of the objective function can be derived as follows:
\begin{align}
    \nabla_\theta J(\theta) &= \mathbb{E}_{{a_{1:T} \sim \pi_\theta}} [\nabla_\theta \log \pi_\theta(a_{1:T}| \es, \relq)R(s_T|\es, \relq)] \\
    &\approx \frac{1}{N} \sum_{n=1}^{N} \nabla_\theta \log \pi_\theta(a^n_{1:T}| \es, \relq)R 
\end{align}
Here, $N$ is the number of policy rollouts.

Each policy rollout explores a sequence of actions $a_{1:T}$. At each timestep $t \in \{1:T\}$, the agent selects an action $a_t$ conditioned on the current state $s_t$. Therefore, the gradient of the log-likelihood in Eq. 6 can be expressed as 
\begin{equation}
    \nabla_\theta \log \pi_\theta(a_{1:T}| \es, \relq) = \sum_{t=1}^{T} \nabla_\theta \log \pi_\theta(a_t| s_t, \es, \relq)
\end{equation}

\paragraph{Reward Shaping.} Previous work \cite{Lin:2018a} observed that a soft reward for the target entities is more beneficial than a binary reward. Following their work, we use pre-trained ConvE \cite{Dettmers:2018} embeddings for the observed entities and relations to shape the reward function. If the agent reaches the correct answer entity, it receives reward $1$. Otherwise, the agent receives a reward estimated by the scoring function of the pre-trained ConvE. Note that the ConvE model is trained only on the training sub-graph of seen entities. ConvE plays no role during inference. Its only purpose is to provide \emph{soft reward} signals during training to help the model in learning a better policy.

\section{Evaluation}
\subsection{Datasets}

\begin{table*}[t!]
    \centering
    \begin{tabular}{lrrrrrrr}
        \hline
        \textbf{Dataset} & $|\mathcal{E}|$ & $|\mathcal{R}|$ & $|\mathcal{U}|$ &  \multicolumn{4}{c}{$|\mathcal{F}|$}\\
        \hline
        &&&& train & dev & test & aux \\
        \cline{5-8}
        FB15k-237-Inductive~~~~~ & 13,119 & ~~237 & 1,389 & ~~227,266 & ~~17,500 & ~~32,197 & ~~61,330\\
        WN18RR-Inductive & 35,928 & 11 & ~~4,029& 67,564 & 3,000 & 11,015 & 19,395\\
        NELL-995-Inductive & 71,578 & 200 & 776 & 137,221 & 500 & 1,679 & 2,267\\
        \hline
    \end{tabular}
    \caption{Evaluation datasets for inductive setting}
    \label{tab:dataset-inductive}
\end{table*}

We evaluate our model based on three standard benchmark knowledge graph completion datasets.
(1) FB15k-237 \cite{Toutanova:2016}, introduced as a replacement for the FB15k dataset \cite{Bordes:2013}. 
In FB15k-237, the reverse relations are removed, rendering the dataset more challenging for inference. (2) WN18RR \cite{Dettmers:2018} is a subset of the WN18 benchmark dataset. Similar to FB15k-237, the reverse relations are removed for this dataset. (3) NELL-995 \cite{Xiong:2017a} is a subset of the 995-th iteration of NELL.

To test the effectiveness of our model for inductive representation learning and reasoning, we create new splits of training, development, and test sets for each of the three benchmark datasets mentioned above.
This new split of the data is necessary, as in an inductive setting, the subject entities in the test set must not be present anywhere in the training subgraph. To satisfy this requirement, we first sample $10\%$ of all the entities present in each of the benchmark datasets. We denote this as the set of \emph{unseen entities} $\mathcal{U}$, while the remaining entities are denoted as \emph{seen entities} $\mathcal{E}$. Then, we proceed to split the triples in the datasets into three disjoint sets. The first set contains the triples in which both the head and the tail entities are in $\mathcal{E}$. The second set consists of the triples with head entities belonging to $\mathcal{U}$, but tail entities in $\mathcal{E}$. In the third set, the head entities belong to  $\mathcal{E}$, but the tail entities are in $\mathcal{U}$. We further split the first set into \emph{train} and \emph{dev} triples. The second set becomes the \emph{test} triples, and the union of the second and the third set is denoted as \emph{auxiliary} data. Auxiliary triples are required to obtain the local neighborhood of a source entity at inference time. Note that an emerging entity in the test set is not disconnected from the training graph. It has at least one seen entity in its neighborhood. This ensures that our model can find a path to the target entity during inference. If the emerging entity were completely disconnected from the training graph (i.e.\ all neighboring nodes were in $\mathcal{U}$), finding a path to the target entity would not be possible.

We append the suffix \emph{"-Inductive"} to distinguish these newly derived datasets from their original counterparts. A summary of these datasets is presented in Table \ref{tab:dataset-inductive}. To help with the reproducibility for future research on this topic, we make the datasets and our source code publicly available at: \\ \texttt{https://github.com/kingsaint/InductiveExplainableLinkPrediction}

\subsection{Baselines}
\paragraph{Embedding Methods.}
We compare our model to a set of embedding based models that perform well under the transductive setting of link prediction. Although these models are particularly unsuitable for the inductive setting, we include them to better demonstrate the challenges of applying such algorithms in an inductive setting. In particular, we compare our model to ConvE \cite{Dettmers:2018}, TransH \cite{TransH:Wang:2014:AAAI}, TransR \cite{TransR:Lin:2015:AAAI}, and RotatE \cite{RotatE:2019:Sun}. For these experiments, we adapted the PyKEEN\footnote{\url{https://github.com/pykeen/pykeen}}
implementations of these models. 

\paragraph{Graph Convolution Methods.}
We choose a state-of-the-art graph convolution-based method CompGCN \cite{CompGCN:2020:Vashishth} as a baseline. Our choice is motivated by two factors: (1) CompGCN performs strongly in the transductive setting by outperforming the other baselines for most of the datasets, and (2) since its encoder module deploys neighborhood integration through Graph Convolution Networks, it has similar characteristics to our model, and therefore, is a good candidate for inductive representation learning. We also compare our model to R-GCN \cite{Schlichtkrull:2018} and SACN \cite{SACN:Shang:2019:AAAI}, which also leverage the graph structure to learn node representations by aggregating neighborhood information. For CompGCN %
and SACN,  
we adapted the source code made available by the authors to make them suitable for inductive representation learning and link prediction. For R-GCN, we adapted the source code available in the DGL library\footnote{\url{https://github.com/dmlc/dgl/tree/master/examples/pytorch/rgcn}}.

\paragraph{Symbolic Rule Mining.}
We compare our model with AnyBURL \cite{AnyBURL:2019:Meilicke}, a purely symbolic rule mining system. AnyBURL is capable of extremely fast rule mining, has outperformed other rule mining approaches including AMIE+ \cite{AMIE+:2015}, and produces comparable results to existing embedding-based models. 

\paragraph{Path-based Model.}
Finally, we compare our model to a policy gradient-based multihop reasoning approach \cite{Lin:2018a} that is similar to the decoder module of our model. We modified the source code\footnote{\url{https://github.com/salesforce/MultiHopKG}} of this model to adapt it to our task. 

\subsection{Experimental Details}
\paragraph{Training Protocol.}
Since the benchmark knowledge graph completion datasets contain only unidirectional edges $(\es, \relq, \eo)$, for all methods, we augment the training sub-graph with the reverse edges $(\eo, r_\mathrm{q}^{-1}, \es)$. During the Graph Transformer based inductive representation learning, $n\%$ of local neighboring entities are randomly selected and masked. During training, we mask $50\%$, $50\%$, and $30\%$ of neighboring nodes, respectively, for the FB15k-237, WN188RR, and NELL-995 datasets. Neighborhood masking helps in learning robust representations and reduces the memory footprint, and has been shown to be effective \cite{GraphSAGE:Hamilton:2017:NIPS}. Following previous work \cite{Das:2017,Lin:2018a}, during training of the policy network, we also retain the top-$k$ outgoing edges for each entity that are ranked by the PageRank scores of the neighboring entities. We set the value of $k$ for each dataset following Lin et al.~\cite{Lin:2018a}. Such a cut-off threshold is necessary to prevent memory overflow. Finally, we adopt the false-negative masking technique in the final timestep of the policy rollouts to guide the agent to the correct answer entities as described in previous work \cite{Das:2017,Lin:2018a}, where it was found helpful when multiple answer entities are present in the training graph.

\paragraph{Hyperparameters.}
For a fair comparison to the baselines, we keep the dimensionality of the entity and relation embeddings at 200. 
For our model, we deploy one layer of a Transformer block ($L = 1$) and $4$ attention heads ($N = 4$). We choose a minibatch size of 64 during training due to limited GPU memory. We rely on Adam \cite{kingma2014method}  stochastic optimization with a fixed learning rate of $0.001$ across all training epochs. Additionally, we adopt entropy regularization to improve the learning dynamics of the policy gradient method. The regularizer is weighted by a hyperparameter $\beta$ set to a value within $[0, 0.1]$. We apply dropout to the entity and relation embeddings, the feedforward networks, and the residual connections. The policy rollout is done for $T = 3$ timesteps for every dataset.

\paragraph{Evaluation Protocol.}
Following previous work \cite{Lin:2018a}, we adopt beam search decoding during inference with a beam width of $512$ for NELL-995 and $256$ for the other datasets. If more than one path leads to the same target entity, then the path with the maximum log-likelihood is chosen over the others. During evaluation, the auxiliary graph augments the training graph to construct the KG environment with unseen entities and their relations to the seen entities. For our model and the baselines, the embeddings of all unseen entities are initialized with Xavier normal initialization \cite{Xavier:PMLR:2010} at inference time.

\paragraph{Evaluation Metrics.} We adopt the ranking based metrics \emph{Mean Reciprocal Rank} and \emph{Hits@k} that are also used by prior work for evaluation. We follow the \emph{filtered setting}  \cite{Bordes:2013} adopted by prior approaches. In the filtered setting, the scores for the false negative answer entities are masked to facilitate correct ranking of the target entity.

\subsection{Results}

\begin{table*}[tp]
    \centering
    \begin{tabular}{lrrrrrrrrrrrrrr}
        \hline
        \multicolumn{1}{c}{} &
        \multicolumn{4}{c}{WN18RR-Inductive} &
        \multicolumn{1}{c}{} &
        \multicolumn{4}{c}{FB15K-237-Inductive} &
        \multicolumn{1}{c}{} &
        \multicolumn{4}{c}{NELL-995-Inductive} \\
        \cline{2-5}\cline{7-10}\cline{12-15}
        & \multicolumn{1}{c}{} &
        \multicolumn{3}{c}{Hits@N} &
        \multicolumn{2}{c}{} &
        \multicolumn{3}{c}{Hits@N} &
        \multicolumn{2}{c}{} &
        \multicolumn{3}{c}{Hits@N} \\
        \cline{3-5}\cline{8-10}\cline{13-15}
        Model & MRR & @1 & @3 & @10 && MRR & @1 & @3 & @10 && MRR & @1 & @3 & @10\\
        \hline
        TransR \cite{TransR:Lin:2015:AAAI} & 0.8 & 0.6 & 0.7 & 0.9 && 5.0 & 4.0 & 5.2 & 6.6 && 5.3 & 4.9 & 5.3 & 6.5 \\
        TransH \cite{TransH:Wang:2014:AAAI} & 0.0 & 0.0 & 0.0 & 0.0 && 6.2 & 5.4 & 6.3 & 8.0 && 3.6 & 3.4 & 3.6 & 3.6 \\
        RotatE \cite{RotatE:2019:Sun} & 0.0 & 0.0 & 0.0 & 0.0 && 0.0 & 0.0 & 0.0 & 0.0 && 0.0 & 0.0 & 0.0 & 0.0 \\
        ConvE \cite{Dettmers:2018} & 1.9 & 1.1 & 2.1 & 3.5 && 26.3 & 20.0 & 28.7 & 38.8 && 43.4 & 32.5 & 50.3 & 60.9\\
        R-GCN \cite{Schlichtkrull:2018} & 14.7 & 11.4 & 15.1 & 20.7 && 19.1 & 11.5 & 20.9 & 34.3 && 58.4 & 50.9 & 62.9 & 71.6\\
        SACN \cite{SACN:Shang:2019:AAAI} & 17.5 & 9.7 & 20.3 & 33.5 && 29.9 & 20.5 & 32.8 & 50.0 && 42.4 & 37.0 & 42.9 & 53.2\\
        CompGCN \cite{CompGCN:2020:Vashishth} & 2.2 & 0.0 & 2.2 & 5.2 && 26.1 & 19.2 & 28.5 & 39.2 && 42.8 & 33.1 & 47.9 & 61.0\\
        AnyBURL \cite{AnyBURL:2019:Meilicke} & - & \textbf{48.3} & 50.9 & 53.9 && - & 28.3 & 43.0 & 56.5 && - & 8.7 & 11.0 & 12.3\\
        MultiHopKG \cite{Lin:2018a} & 45.5 & 39.4 & 49.2 & 56.5 && 38.6 & 29.3 & 43.4 & 56.7 && 74.7 & 69.1 & 78.3 & 84.2 \\
        \hline
        Our Model w/ RS & \textbf{48.8} & 42.1 & \textbf{52.2} & \textbf{60.6} && \textbf{39.8} & \textbf{30.7} & \textbf{44.5} & \textbf{57.6} && \textbf{75.2} & \textbf{69.7} & \textbf{79.1} & \textbf{84.4}\\
        \hline
    \end{tabular}
    \caption{Evaluation results of our model as compared to 
    alternative baselines on inductive variants of the
    WN18RR, FB15K-237, and NELL-995 datasets.
    The Hits@N and MRR metrics are multiplied by 100.}
    \label{tab:results}
\end{table*}

We present the experimental results of our method and the baselines in Table \ref{tab:results}. The results of the embedding-based models TransH, TransR, and RotatE across all datasets demonstrates their inability to deal with entities that are unseen during training. These models are thus rendered as ineffective for inductive representation learning and reasoning. ConvE performs better than other embedding-based models we consider. Still, the much inferior performance of ConvE compared to our model shows that ConvE is not particularly suitable for inductive representation learning.

We observe that our model significantly outperforms the graph convolution network baselines CompGCN, SACN, and R-GCN across all datasets. Although these models use the neighborhood information for learning representations, unlike our method, their neighborhood integration methods do not explicitly consider the query relations. 

We find AnyBURL and MultiHopKG to be the most competitive methods to ours. AnyBURL performs adequately for the WN18RR and FB15K-237 dataset while performing poorly on the NELL-995 dataset. MultiHopKG adapts surprisingly well to our dataset despite the unseen entities being initialized with Xavier normal initialization. We conjecture that the learned representations of the query and the outgoing edges (relations) have enough semantic information encoded in them to navigate to the target entity by simply exploiting the edge (relation) information. However, our proposed model holds an edge over this model with $7.2\%$, $3.1\%$, and $0.7\%$ gains in the MRR metric for the WN18RR, FB15K-237, and NELL-995 datasets respectively.

\section{Analysis}
In this section, we perform further analysis of our proposed model. First, we conduct a set of ablation studies. Then, we qualitatively analyze our model's ability to provide reasoning paths as supporting evidence for inference. Finally, we analyze the effect of the cardinality of relation types on the inference process.

\subsection{Ablation Study}
\begin{table*}[tp]
    \centering
    \begin{tabular}{lrrrrrrrrrrrrrr}
        \hline
        \multicolumn{1}{c}{} &
        \multicolumn{4}{c}{WN18RR-Inductive} &
        \multicolumn{1}{c}{} &
        \multicolumn{4}{c}{FB15K-237-Inductive} &
        \multicolumn{1}{c}{} &
        \multicolumn{4}{c}{NELL-995-Inductive} \\
        \cline{2-5}\cline{7-10}\cline{12-15}
        \multicolumn{2}{c}{} &
        \multicolumn{3}{c}{Hits@N} &
        \multicolumn{2}{c}{} &
        \multicolumn{3}{c}{Hits@N} &
        \multicolumn{2}{c}{} &
        \multicolumn{3}{c}{Hits@N} \\
        \cline{3-5}\cline{8-10}\cline{13-15}
        Model & MRR & @1 & @3 & @10 && MRR & @1 & @3 & @10 && MRR & @1 & @3 & @10\\
        \hline
        Our Model w/ RS & \textbf{48.8} & 42.1 & \textbf{52.2} & \textbf{60.6} && \textbf{39.8} & \textbf{30.7} & \textbf{44.5} & \textbf{57.6} && \textbf{75.2} & \textbf{69.7} & \textbf{79.1} & \textbf{84.4}\\
        Our Model w/o RS & 48.2  & 40.1 & 53.0 &  62.2 && 37.8 & 29.4 & 42.6 & 54.0 && 71.1 & 65.3 & 75.0 & 79.9 \\
        GT + ConvTransE & 1.1 & 0.6 & 1.1 & 1.8 && 22.9 & 17.3 & 24.9 & 33.3 && 47.9 & 40.6 & 51.1 & 61.9 \\
        \hline
    \end{tabular}
    \caption{Ablation study. The Hits@N and MRR metrics are multiplied by 100.}
    \label{tab:results_4}
\end{table*}

To better understand the contribution of reward shaping in our model, we perform an ablation study, where our model is deprived of the \emph{soft reward} signals provided by ConvE. In general, we observe that replacing reward shaping  with hard binary reward deteriorates the performance of our model across all datasets. Note that our ablated version still mostly outperforms the other baseline methods. 

Additionally, we experiment with a non-explainable variant of our model, in which we retain the Graph Transformer (GT) encoder, but we replace the policy gradient-based decoder with an embedding-based decoder called ConvTransE, which is also used in SACN as a decoder module. With this model, we observe a significant drop in performance. Thus, we conjecture that the policy gradient-based decoder not only provides explainability, but also is crucial for decoding.

\subsection{Qualitative Analysis of Explainability}
Since explainability is one of the key objectives of our model, we provide examples of explainable reasoning paths for queries that involve previously unseen source entities at inference time. Table \ref{tab:path_examples} contains examples of $1$-hop, $2$-hop, and $3$-hop reasoning paths. These examples demonstrate our model's effectiveness in learning inductive representations for the unseen entities, which helps to infer the reasoning paths.

\subsection{Effect of Relation Types}

\begin{table*}[tp]
    \centering
    \begin{tabular}{lp{0.8\textwidth}}
        \hline
        \textbf{Query} & (William Green, worksFor, ?) \\
        \textbf{Answer} & Accenture \\
        \textbf{Explanation~~} & William Green $\xrightarrow{\mathrm{personLeadsOrganization}}$ Accenture \\
        
        \hline
        \textbf{Query} & (Florida State, organizationHiredPerson, ?)\\ 
        \textbf{Answer} & Bobby Bowden \\
        \textbf{Explanation} & Florida State $\xleftarrow{\mathrm{worksFor}}$ Bobby Bowden\\
        
        \hline 
        \textbf{Query} & (Messi, athleteHomeStadium, ?) \\
        \textbf{Answer} & Camp Nou  \\
        \textbf{Explanation} & Messi $\xrightarrow{\mathrm{athletePlaysForTeam}}$ Barcelona $\xrightarrow{\mathrm{teamHomeStadium}}$  Camp Nou
        \\
        \hline
        
        \textbf{Query} & (Adrian Griffin, athleteHomeStadium, ?) \\
        \textbf{Answer} & United Center \\
        \textbf{Explanation} & Adrian Griffin $\xrightarrow{\mathrm{athletePlaysForTeam}}$  Knicks   $\xleftarrow{\mathrm{athletePlaysForTeam}}$ Eddy Curry $\xrightarrow{\mathrm{athleteHomeStadium}}$ United Center \\
        \hline
        
         \textbf{Query} & (Bucks, teamPlaysInLeague, ?) \\
        \textbf{Answer} & NBA \\
        \textbf{Explanation} & Bucks $\xrightarrow{\mathrm{organization
HiredPerson}}$  Scott Stiles   $\xleftarrow{\mathrm{organizationHiredPerson}}$ Chicago Bulls $\xrightarrow{\mathrm{teamPlaysInLeague}}$ NBA \\
        \hline
    \end{tabular}
    \caption{Example queries from the NELL-995 test set with unseen source entities. The answers are supported by the explainable reasoning paths derived by our model.}
    \label{tab:path_examples}
\end{table*}
\begin{table}[!h]
    \centering
    \begin{tabular}{lcc|cc}
    \hline
        \multicolumn{1}{l}{Dataset} & \multicolumn{2}{c}{to-Many} & \multicolumn{2}{c}{to-1} \\
        \hline
        & \% & MRR & \% & MRR \\
        \cline{2-5}
        FB15k-237-Inductive~~~~ & 77.4 & 31.6 & 22.6 & 75.5\\
        WN18RR-Inductive & 48.1 & 60.8 & 51.9 & 30.1\\
        NELL-995-Inductive & 7.6 & 41.4 & 92.4 & 78.5\\
    \hline
    \end{tabular}
    \caption{MRR for the test triples in inductive setting with \emph{to-Many} and \emph{to-1} relation types. The \% columns show the percentage of test triples for each relation type.}
    \label{tab:relation_type}
\end{table}

Following Bordes et al. \cite{Bordes:2013}, we categorize the relations in the seen snapshot of the knowledge graph into Many-to-1 and 1-to-Many relations. The categorization is done based on the ratio of the cardinality of the target answer entities to the source entities. If the ratio is greater than $1.5$, we categorize the relation as \emph{to-Many}, otherwise as \emph{to-1}.
We analyzed the results of the test set for these two types of relations. We report the percentage of triples with these two types of relations and the corresponding MRR achieved by our model in Table \ref{tab:relation_type}. For FB15k-237 and NELL-995, our model performs better for \emph{to-1} relations than for \emph{to-many} relations. On the contrary, we observe a reverse trend for the WN18RR dataset. Note however that \emph{to-many} relations have alternative target entities. In the current evaluation protocol, our model is punished for predicting any alternative target entity other than the ground truth target.

\section{Conclusion}
The ever-expanding number of entities in knowledge graphs warrants the exploration of knowledge graph completion methods that can be applied to emerging entities without retraining the model. 
While prior approaches assume a static snapshot of the knowledge graph, we introduce a joint framework for inductive representation learning to predict missing links in a dynamic knowledge graph with many emerging entities. Additionally, our method provides explainable reasoning paths for the inferred links as support evidence. Through experiments we demonstrate that our model significantly outperforms the baselines across the new inductive benchmark datasets introduced in this paper.

\section{Acknowledgement}
We thank Diffbot for their grant support to Rajarshi Bhowmik's work. We also thank Diffbot and Google for providing the computing infrastructure required for this project.

\bibliographystyle{splncs04}
\bibliography{bibliography}

\end{document}